\pdfoutput=1

\documentclass[11pt]{article}

\usepackage[]{acl}

\usepackage{times}
\usepackage{latexsym}
\usepackage{amssymb}

\usepackage[T1]{fontenc}

\usepackage[utf8]{inputenc}

\usepackage{microtype}

\usepackage{inconsolata}
\usepackage{enumitem}
\usepackage{algorithm}
\usepackage{microtype}
\usepackage{boldline}
\usepackage{multirow}
\usepackage{makecell}
\usepackage{hhline}
\usepackage{pgfplots}
\usepackage{tikz}
\usepackage{enumitem}
\usepackage{colortbl}
\usepackage{mathtools}
\usepackage{pifont}
\usepackage{bbm}
\usepackage{amsfonts}
\usepackage{algpseudocode}
\usepackage{booktabs}
\usepackage{xparse}
\usepackage{graphicx}
\usepackage{multicol}
\usetikzlibrary{shapes,arrows}
\usepackage{natbib}
\usepackage{caption}
\usepackage{subcaption}
\usepackage{tikz-dependency}
\usepackage{wrapfig}
\pgfplotsset{compat=1.17} 
\usepackage{subfiles}
\usepackage{readarray}
\usepackage{xcolor}
\usepackage{import}

\newcommand{\mcL}{\mathcal{L}}

\usepackage{amsmath,amsfonts,bm}

\def\eqref#1{equation~\ref{#1}}

\def\1{\bm{1}}

\def\vy{{\bm{y}}}

\DeclareMathAlphabet{\mathsfit}{\encodingdefault}{\sfdefault}{m}{sl}
\SetMathAlphabet{\mathsfit}{bold}{\encodingdefault}{\sfdefault}{bx}{n}

\usepackage{amsmath}

\title{LLMs as Bridges: Reformulating Grounded Multimodal Named Entity Recognition}

\author{Jinyuan Li$^1$, Han Li$^2$, Di Sun$^4$, Jiahao Wang$^3$, Wenkun Zhang$^5$, Zan Wang$^{1,3}$, \textbf{Gang Pan$^1$$^{,3,}$\thanks{\hspace{1mm} corresponding authors. }}\\
 $^1$School of New Media and Communication, Tianjin University \\
 $^2$College of Mathematics, Taiyuan University of Technology \\
 $^3$College of Intelligence and Computing, Tianjin University \\
 $^4$Tianjin University of Science and Technology\\ 
 $^5$University of Copenhagen \\
  {\tt \{jinyuanli, wjhwtt, wangzan, pangang\}@tju.edu.cn,} 
  {\tt lihan0928@link.tyut.edu.com} \\
  {\tt wenkun.zhang@sund.ku.dk,}
  {\tt dsun@tust.edu.cn} \\
}

\begin{document}
\maketitle
\begin{abstract}

Grounded Multimodal Named Entity Recognition (GMNER) is a nascent multimodal task that aims to identify named entities, entity types and their corresponding visual regions. GMNER task exhibits two challenging properties: 1) The weak correlation between image-text pairs in social media results in a significant portion of named entities being ungroundable. 2) There exists a distinction between coarse-grained referring expressions commonly used in similar tasks (\emph{e.g.}, phrase localization, referring expression comprehension) and fine-grained named entities. In this paper, we propose RiVEG, a unified framework that reformulates GMNE\underline{R} \underline{i}nto a joint MNER-\underline{VE}-V\underline{G} task by leveraging large language models (LLMs) as a connecting bridge. This reformulation brings two benefits: 1) It maintains the optimal MNER performance and eliminates the need for employing object detection methods to pre-extract regional features, thereby naturally addressing two major limitations of existing GMNER methods. 2) The introduction of entity expansion expression and Visual Entailment (VE) module unifies Visual Grounding (VG) and Entity Grounding (EG). It enables RiVEG to effortlessly inherit the Visual Entailment and Visual Grounding capabilities of any current or prospective multimodal pretraining models. Extensive experiments demonstrate that RiVEG outperforms state-of-the-art methods on the existing GMNER dataset and achieves absolute leads of 10.65\%, 6.21\%, and 8.83\% in all three subtasks. Code and datasets publicly available at \url{https://github.com/JinYuanLi0012/RiVEG}.

\end{abstract}

\section{Introduction}
\begin{figure}[h]
  \scriptsize
      \setlength{\belowcaptionskip}{-0.5cm}
    \centering  \includegraphics[height=2.05in,width=3.1in]{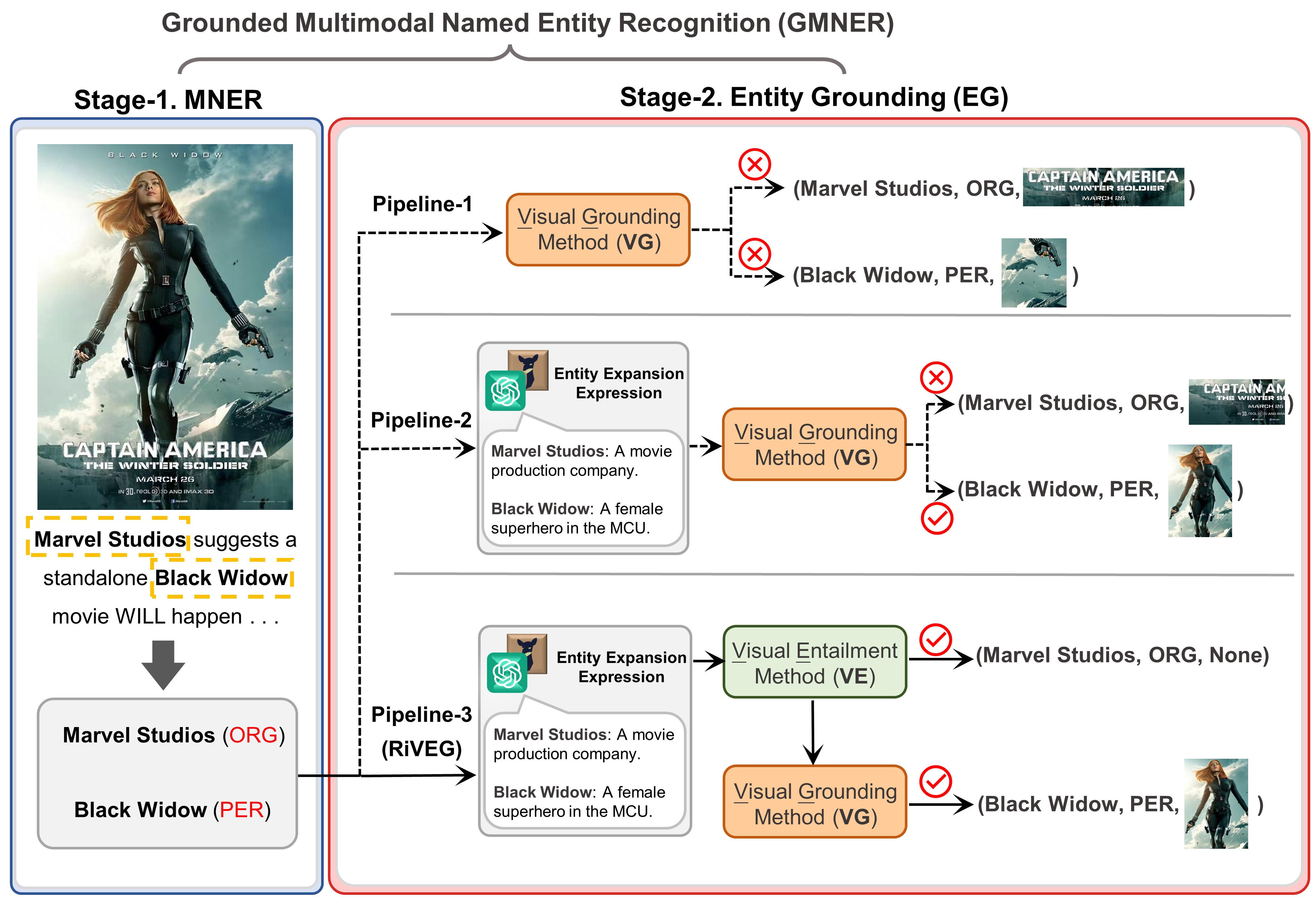}
  \caption{Pipeline-1 illustrates the challenges posed by GMNER task to existing Visual Grounding methods. RiVEG builds a bridge between noun phrases and named entities and introduces Visual Entailment to address the weak correlation between image and text.}
  \label{headphoto}
\end{figure}
Multimodal Named Entity Recognition (MNER) \citep{lu2018visual} aims to extract named entities and corresponding categories from image-text pairs sourced from social media. As one of the most important subtask in multimodal information processing, MNER has received extensive attention in recent years \citep{wang-etal-2022-ita, wang-etal-2022-named, li-etal-2023-prompting, cui2024enhancing}. Due to the pressing demand for constructing structured knowledge graphs, human-machine interaction and robotics, recent research endeavors to extend the scope of MNER \citep{wang2023fine}. 
As a further extension, Grounded Multimodal Named Entity Recognition (GMNER) \citep{yu-etal-2023-grounded} aims to extract text named entities, entity types and their bounding box groundings from image-text pairs.

Analyzing GMNER from the two perspectives of MNER and EG, existing GMNER methods \citep{yu-etal-2023-grounded} have several significant limitations. 
Limitation 1. Entity Grounding (EG) is an important part in GMNER, which aims to determine the groundability and visual grounding results of text named entities predicted by the MNER model. 
Obviously, effective EG depends on accurate MNER. 
But the MNER performance of the existing multimodal fusion GMNER methods \citep{yu-etal-2023-grounded} is suboptimal. 
This may be due to the introduction of visual object features destroying the original text feature representation in MNER \citep{wang-etal-2022-ita}. A series of MNER studies in the Text-Text paradigm (\emph{i.e.}, represent images using text) \citep{wang-etal-2022-named, li-etal-2023-prompting} show that text representation plays the most critical role in MNER. Models with the best MNER performance typically do not involve any cross-modal interactions. However, existing end-to-end GMNER methods \citep{yu-etal-2023-grounded} violate this in order to obtain EG capabilities with the help of cross-modal interactions. This results in the feature representation learned by the model being neither the best for MNER nor the best for EG. Therefore, splitting the prediction and grounding of named entities into separate stages is necessary. Limitation 2. Existing GMNER methods \citep{yu-etal-2023-grounded, wang2023fine} typically employ object detection (OD)  techniques to extract features from candidate regions, which are then aligned with text entities. But as shown in Table \ref{tab:OD test}, our statistical results\footnote{Here, TopN-Prec@0.5 refers to the accuracy metric in object detection methods across all samples. It measures the proportion of named entities where at least one of the Top-N predicted bounding boxes based on detection probability has an IoU of 0.5 or greater with the ground-truth bounding box.} indicate that the widely-adopted OD methods exhibit unsatisfactory performance in recognizing candidate visual objects in GMNER dataset. This implies that there exists a natural upper limit to the effectiveness of existing GMNER methods. In instances where the OD methods are unable to detect the visual ground-truth bounding boxes for textual named entities, even if the MNER part predicts the correct textual named entity, the EG result will definitely be wrong. Therefore, it is necessary to avoid extracting visual features from candidate image regions.

Visual Grounding (VG) \citep{rohrbach2016grounding, yu2018rethinking} aims to correlate natural language descriptions with specific regions in images. Although VG has a similar paradigm to the Entity Grounding (EG) in GMNER, existing VG methods \citep{zhu2022seqtr, deng2021transvg, wang2022ofa} exhibit significant deficiencies when applied to the GMNER dataset. This is attributed to two fundamental distinctions between these two tasks: 
Differentiation 1. Referring to the Pipeline-1 and Pipeline-2 in Figure \ref{headphoto}, the text input for the EG task consists of real-world named entities, which is notably different from the referring expressions or noun phrases typically used in the VG method. There is a substantial difference that is almost non-generalizable between them. The VG method can comprehend the description ``A female superhero in the MCU'' but struggles with understanding named entities like ``Black Widow''. Therefore, a bridge between noun phrases and named entities needs to be built. 
\begin{table}[t!]
\small
\setlength\tabcolsep{6.3pt}
\renewcommand{\arraystretch}{1.2}
\centering
\begin{tabular}{l|ccc}
\toprule
OD methods & Top-10 & Top-15 & Top-20  \\
\midrule
\citet{anderson2018bottom} & 59.87\% & 69.84\% & 76.11\%  \\
\citet{zhang2021vinvl} & 66.69\%  & 74.62\% & 84.29\%   \\
\bottomrule
\end{tabular}
\caption{TopN-Prec@0.5 scores of two widely-adopted OD methods on Twitter-GMNER dataset \citep{yu-etal-2023-grounded}. 
Existing GMNER methods are significantly limited by OD techniques. Because these visual candidate areas do not necessarily include the visual gold label of Entity Grounding.}
\label{tab:OD test}
\end{table}
Differentiation 2. Due to the social media usage patterns of users, image-text pairs from social media exhibit apparent weak correlation. It implies that the textual named entity may not necessarily be linked to a specific region in the image. \citet{yu-etal-2023-grounded} shows that approximately 60\% of the named entities in the GMNER dataset are ungroundable. But the classic VG task stipulates that the input referring expression must match an object in the image, without considering ungroundable expressions that do not match any object. It means that the behavior of the existing VG methods is undefined if the named entity does not correspond to any object in the image. As shown in Pipeline-1 and Pipeline-2 of Figure \ref{headphoto}, existing VG methods inevitably make errors when they confront the ungroundable text input like ``Marvel Studios''. Our experiments demonstrate that the Prec@0.5 of the state-of-the-art VG methods applied to the GMNER dataset is 21.87\% \citep{wang2022ofa} and 9.23\% \citep{zhu2022seqtr}, respectively.\footnote{We fine-tune VG models using the GMNER dataset. Text inputs are named entities and entity types. For ungroundable entities, the coordinates of gold labels are defined as all zeros.} Therefore, a separate module is needed to determine the groundability of text inputs.

Given the above differences, the potential of VG methods cannot be unleashed in EG. Considering that most VG methods do not require pre-extraction of regional features, we ask: \emph{Is it possible to reformulate GMNER as a two-stage union of MNER-EG? The first stage focuses on optimizing the MNER effect to solve the Limitation 1 of the existing GMNER method. The second stage uses the VG method to naturally bypass regional feature extraction to solve Limitation 2. In the second stage, to adapt the VG method to the EG task, is it conceivable to establish a bridge between named entities and noun phrases to address Differentiation 1 between the VG method and EG task? Moreover, is it feasible to resolve Differentiation 2 by devising a sensible filtering mechanism?}

In this paper, we introduce GMNE\underline{R} \underline{i}nto a multi-stage framework consisting of MNER, \underline{VE} and V\underline{G} (RiVEG). This modular design inherently mitigates two limitations of existing GMNER methods, enabling the connection of three unrelated independent tasks in series within GMNER and unlocking significant potential. Specifically, in order to ensure the optimal performance of the MNER stage, we leverage auxiliary refined knowledge distilled from large language models (LLMs) to aid in the identification of named entities. And we introduce a Visual Entailment (VE) module to handle the weak correlation between image and text. Named entities are heuristically converted by LLMs into corresponding entity expansion expressions to facilitate the VE and VG modules to determine the groundability and grounding results of named entities. Main contributions are summarized as follows:
\begin{enumerate}[label=(\roman*)]
    \vspace{-3pt}
    \item
    RiVEG extends the text input of Visual Entailment and Visual Grounding from limited natural language expressions that require manual definition to infinite named entities that exist naturally by leveraging LLMs as bridges. 
    \vspace{-3pt}
    \item
    RiVEG increases the application scope of VG models to the cases of no right objects in image and demonstrates a new paradigm for GMNER. This paradigm supports data augmentation and facilitates subsequent upgrades. 
    \vspace{-3pt}
    \item
    All 14 variants of RiVEG achieve new state-of-the-art performance on the GMNER dataset. And after effective data augmentation, the MNER module of RiVEG achieves new state-of-the-art performance on two classic MNER datasets. 
\end{enumerate}
\vspace{-3pt}
    
\section{Related Work}
\subsection{Multimodal Named Entity Recognition}
MNER is a relatively independent multimodal subtask. Early MNER methods attempt to simply interact text and images \citep{moon-etal-2018-multimodal, lu-etal-2018-visual, yu-etal-2020-improving-multimodal}. To deal with the limitations of this shallow interaction, subsequent methods explore various cross-modal fusion approaches \citep{zhang2021multi, wang2022cat, jia2023mner, chen2022hybrid, chen2022good}. There is a trend to solve MNER through the Text-Text paradigm \citep{wang-etal-2022-ita, wang-etal-2022-named, li-etal-2023-prompting}, and this paradigm generally yields better performance. Despite obtaining promising results, these methods lack any visual understanding capabilities. The GMNER task poses significant challenges to the above research. 

\begin{figure*}[t!]
	\centering
	\includegraphics[scale=0.48]{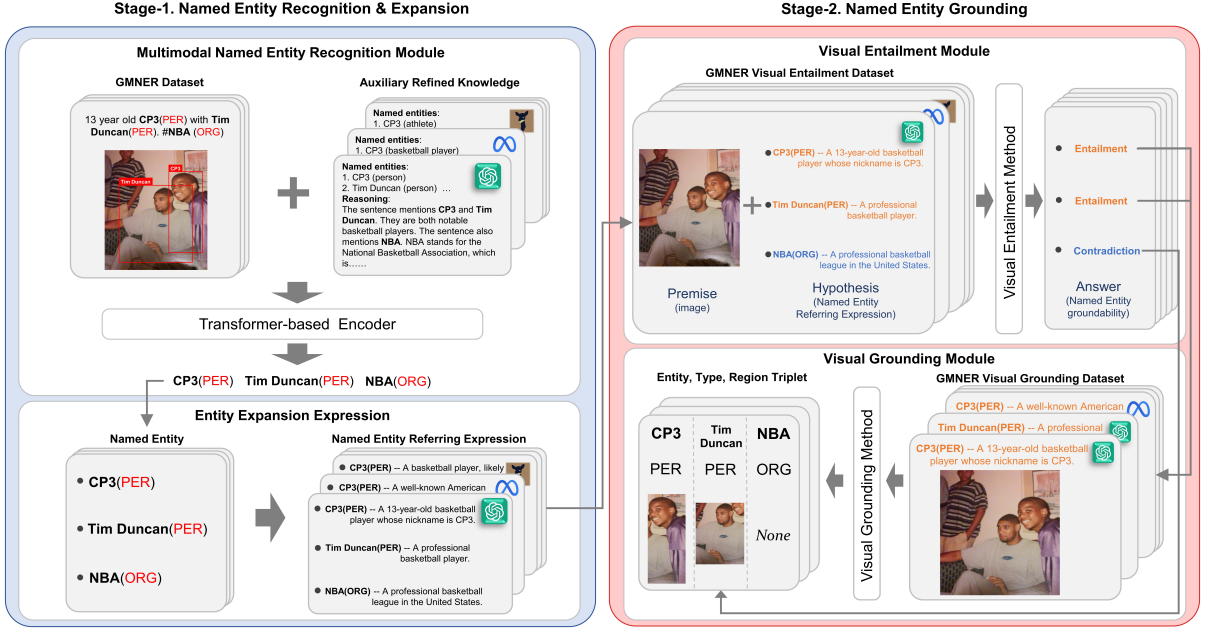}
	\caption{The architecture of RiVEG.}
	\label{fig:architecture}
\end{figure*}

\subsection{Grounded Multimodal Named Entity Recognition and Visual Grounding}
GMNER is a further exploration of MNER. Existing GMNER works \citep{yu-etal-2023-grounded, wang2023fine} aim to utilize OD methods for obtaining region proposals and subsequently sorting them based on their relevance to named entities. Such methods are inevitably limited by OD methods. Visual Grounding (VG) is a representative multimodal subtask. The progress in multimodal pretraining research \citep{zhu2022seqtr, wang2022ofa, yan2023universal} continues to contribute to the improvement of VG methods. However, the VG task has some strong predefined constraints. It presupposes that there are objects in the image that match the input text query. This renders VG methods inapplicable to the GMNER task that deviates from such predefined constraints. In this paper, we introduce the VE module to reconcile this conflict. Our method uses the VG method as the main body of visual understanding and achieves better performance than the existing GMNER methods.

\section{Methodology}
RiVEG is mainly divided into two stages. In the stage of Named Entity Recognition and Expansion, we employ auxiliary refined knowledge obtained from LLMs to ensure optimal MNER performance. Furthermore, in order to adapt existing VG methods to EG task, LLMs is considered as a bridge between named entities and referring expressions. RiVEG instructs LLMs to convert named entities into named entity referring expressions (detailed in §\ref{sec:2.2}). In the stage of Named Entity Grounding, RiVEG reformulates the entire EG task as a union of VE and VG. To reconcile the challenge between the weak correlation of image-text pairs in GMNER and the strong predefined constraints of existing VG methods, we design the VE module for buffering. Images and named entity referring expressions serve as the input to VE module for determining the groundability of the named entities. Finally, the named entities deemed groundable alongside images are employed as inputs to VG module to obtain the final grounding results (detailed in §\ref{sec:2.3}). We also design a data augmentation method suitable for RiVEG. An overview of the RiVEG is depicted in Figure \ref{fig:architecture}.

\subsection{Task Formulation}
Given a sentence $S = \{s_1, \cdots, s_n\}$ with $n$ tokens and its corresponding image $I$, the objective of GMNER is to identify and categorize named entities (\emph{i.e.}, $PER$, $LOC$, $ORG$ and $MISC$) within the sentence while concurrently determining the visually grounded region corresponding to each entity. Formulate that the output of GMNER comprises a set of multimodal entity triples: 
\begin{equation}
	Y = \{(e_1,t_1,v_1), \cdots, (e_n,t_n,v_n)\} \nonumber
\end{equation}
where $e_i$ is one of the entities in sentence, $t_i$ refers to the type of $e_i$, and $v_i$ represents the corresponding visually grounded region. Note that if $e_i$ is ungroundable, $v_i$ is specified as $None$; otherwise, $v_i$ consists of a 4D coordinates representing the top-left and bottom-right locations of the grounded bounding box, \emph{i.e.}, ($v_i^{x_1}, v_i^{y_1}, v_i^{x_2}, v_i^{y_2}$).

\subsection{Stage-1. Named Entity Recognition and Expansion}\label{sec:2.2}
\paragraph{Multimodal Named Entity Recognition Module} 
Incorporating document-level extended knowledge into original multimodal samples has been proven to be one of the most effective methods for MNER \citep{wang-etal-2022-ita, wang-etal-2022-named, li-etal-2023-prompting}. PGIM \citep{li-etal-2023-prompting} points out that despite the inherent limitations of generative models in sequence labeling tasks, LLM serve as an implicit knowledge base that is highly suitable for MNER task. We use the auxiliary refined knowledge\footnote{Original text and image captions are used as inputs to LLMs, and LLMs are guided to generate extended explanations of the original samples.} provided by PGIM to enhance the predictive performance of named entities. And we additionally use the same algorithm to generate different auxiliary refined knowledge for the same sample from more versions of LLMs to perform data augmentation. Specifically, definite auxiliary refined knowledge retrieved by LLMs as $Z = \{z_1, \cdots, z_m\}$. We take the concatenation of the original sentence $S = \{s_1, \cdots, s_n\}$ together with the auxiliary refined knowledge $Z$ as the input of the transformer-based encoder:
\vspace{-3pt}
\begin{displaymath}
 \text{embed} ([S;Z]) =
\{r_1,\cdots,r_n,\cdots,r_{n+m}\} 
\vspace{-3pt}
\end{displaymath}
And the acquired token representations $R =\{r_1,\cdots,r_n\}$ is fed into a linear-chain Conditional Random Field (CRF) \citep{lafferty2001conditional} to generate the predicted sequence $y = \{y_1, \cdots, y_n\}$: 
\vspace{-3pt}
\begin{align}
    P(y|S,Z) &= \frac{\prod\limits_{i=1}^{n} \psi(y_{i-1}, y_i, r_i)}{\sum\limits_{\vy' \in Y} \prod\limits_{i=1}^{n} \psi(y'_{i-1}, y'_i, r_i)}\nonumber
\end{align}
where $\psi$ is the potential function 
and $Y$ is the set of all possible label sequences given the input $S$ and $Z$. Finally, given the input sequence with gold labels $\vy^*$, 
we employ the negative log-likelihood (NLL) loss function for training model: 
\vspace{-3pt}
\begin{align}
\mcL_{\text{MNER}}(\theta) = - \log P_{\theta}(\vy^*|S,Z) \label{eq:nll_loss}\nonumber
\vspace{-3pt}
\end{align}
Note that the approach employed by this module is not unique and can be replaced or upgraded by any more powerful MNER methods in the future. 

\paragraph{Entity Expansion Expression} 
After extracting named entities in the sentence, RiVEG further guides LLMs to reformulate the fine-grained named entities into coarse-grained entity expansion expressions. Specifically, for each extracted named entity, we formulate it as the following template: 
\begin{table}[H]
\vspace{-6pt}
    \fontsize{9}{9}\selectfont
    \centerline{
    \begin{tabular}{|l|}
    \hline
    \textbf{Background}: \textcolor{blue}{$\{Image\ \allowbreak Caption\}$} \\
    \textbf{Text}: \textcolor{blue}{$\{Sentence\}$} \\
    \textbf{Question}: \texttt{In the context of the provided }\\
    \texttt{information, tell me briefly what is}\\
    \texttt{the \textcolor{blue}{$\{Named\ \allowbreak  Entity\}$} in the Text?} \\
    \textbf{Answer}: \textcolor{blue}{$\{Entity\ \allowbreak Expansion\ \allowbreak  Expression\}$} \\
    \hline
    \end{tabular} {}}
    \vspace{-6pt}
\end{table}
\noindent
where \textcolor{blue}{$\{Entity\ \allowbreak Expansion\ \allowbreak Expression\}$} is reserved for LLMs to generate. A limited number of predefined in-context examples are employed to guide the format of responses (detailed in Appendix Figure \ref{in-context sample}). Finally, we concatenate named entities, entity categories, and entity expansion expressions into the final named entity referring expressions: $Named\ \allowbreak Entity(Entity\ \allowbreak Categoriy)-Entity\ \allowbreak Expansion\ \allowbreak Expression$

Furthermore, in order to perform data augmentation, we use different LLMs to generate different entity expansion expressions of the same entity.

\subsection{Stage-2. Named Entity Grounding} \label{sec:2.3}
Through the aforementioned transformation, the GMNER task is naturally reframed as the MNER-VE-VG task. Notably, a large number of well-established studies exist on MNER, VE, and VG. The VE and VG methods that can be utilized at this stage are also not unique. 

\paragraph{Visual Entailment Module}
For existing VG methods, fine-tuning with limited data hardly equips them with the ability to effectively handle irrelevant text and images. Hence, addressing the weak correlation between images and text in GMNER poses the primary challenge in integrating VG methods into GMNER. To achieve this objective, RiVEG introduces the VE module before the VG module. 

For the classic VE task, denote the VE dataset $\mathcal{D_{\text{VE}}}$ as: 
\vspace{-3pt}
\begin{equation}
    \mathcal{D_{\text{VE}}}=\{(i_1, h_1, l_1), \cdots, (i_n, h_n, l_n)\} \nonumber
    \vspace{-3pt}
\end{equation}
where $i_i$, $h_i$, $l_i$ represent an image premise, a text hypothesis and a class label, respectively. Three labels $e$, $n$ or $c$ are determined by the relationship conveyed by $(i_i, h_i)$. Specifically, $e$ (entailment) represents $i_i \vDash h_i$, $n$ (neutral) represents $i_i \nvDash h_i \land i_i \nvDash \neg h_i$, $c$ (contradiction) represents $i_i \vDash \neg h_i$. Similar but distinct, RiVEG defines the GMNER dataset $\mathcal{D_{\text{GMNER(VE)}}}$ as: 
\vspace{-3pt}
\begin{equation}
    \mathcal{D_{\text{GMNER(VE)}}}=\{(i_1, h_1, l_1), \cdots, (i_m, h_m, l_m)\} \nonumber
    \vspace{-3pt}
\end{equation}
where $i_i$, $h_i$, $l_i$ represent an image that corresponds to a named entity, a named entity referring expression and a class label, respectively. Note that since the groundability of a named entity in GMNER dataset is unambiguous, only two labels $e$ and $c$ are retained. $e$ represents $h_i$ is groundable and $c$ represents $h_i$ is ungroundable. Finally, RiVEG replaces the original VE dataset with $\mathcal{D_{\text{GMNER(VE)}}}$ to fine-tune the existing vanilla VE model. In the inference phase, only the samples deemed groundable by the fine-tuned vanilla VE model will be fed into the final VG module. For the remaining samples that are judged to be ungroundable, their entity grounding results are directly defined as $None$. 

\begin{table*}[!]
\small
\setlength\tabcolsep{3.2pt}
\renewcommand{\arraystretch}{1.1}
\centering
\begin{tabular}{lcccccccccc}
\toprule
\multirow{2}{*}{\textbf{Methods}}
 & \multicolumn{3}{|c|}{\textbf{GMNER}}  & \multicolumn{3}{c|}{\textbf{MNER}} & \multicolumn{3}{c}{\textbf{EEG}}  \\
\cline{2-10}
 & \multicolumn{1}{|c}{\textbf{Pre.}} & \textbf{Rec.} & \multicolumn{1}{c|}{\textbf{F1}} & \textbf{Pre.} & \textbf{Rec.} & \multicolumn{1}{c|}{\textbf{F1}} & \textbf{Pre.} & \textbf{Rec.} & \multicolumn{1}{c}{\textbf{F1}} \\
 \midrule 
 \multicolumn{10}{c}{ Text}\\
 \midrule
 HBiLSTM-CRF-None \citep{lu2018visual} & \multicolumn{1}{|c}{43.56} & 40.69 & \multicolumn{1}{c|}{42.07} & 78.80 & 72.61 & 75.58 & \multicolumn{1}{|c}{49.17} & 45.92 & 47.49 \\
 BERT-None \citep{devlin-etal-2019-bert} & \multicolumn{1}{|c}{42.18} & 43.76 & \multicolumn{1}{c|}{42.96} & 77.26 & 77.41 & 77.30 & \multicolumn{1}{|c}{46.76} & 48.52 & 47.63 \\
 BERT-CRF-None \citep{yu-etal-2023-grounded} & \multicolumn{1}{|c}{42.73} & 44.88 & \multicolumn{1}{c|}{43.78} & 77.23 & 78.64 & 77.93 & \multicolumn{1}{|c}{46.92} & 49.28 & 48.07 \\
 BARTNER-None \citep{yan-etal-2021-unified-generative} & \multicolumn{1}{|c}{44.61} & 45.04 & \multicolumn{1}{c|}{44.82} & 79.67 & 79.98 & 79.83 & \multicolumn{1}{|c}{48.77} & 49.23 & 48.99 \\
  \midrule 
 \multicolumn{10}{c}{ Text+Image}\\
 \midrule
  GVATT-RCNN-EVG \citep{lu2018visual} & \multicolumn{1}{|c}{49.36} & 47.80 & \multicolumn{1}{c|}{48.57} & 78.21& 74.39& 76.26 & \multicolumn{1}{|c}{54.19} & 52.48 & 53.32 \\
  UMT-RCNN-EVG \citep{yu-etal-2020-improving-multimodal}  & \multicolumn{1}{|c}{49.16} & 51.48 & \multicolumn{1}{c|}{50.29} & 77.89 &79.28 &78.58 & \multicolumn{1}{|c}{53.55} & 56.08 & 54.78 \\
  UMT-VinVL-EVG \citep{yu-etal-2020-improving-multimodal} & \multicolumn{1}{|c}{50.15} & 52.52 & \multicolumn{1}{c|}{51.31} & 77.89 &79.28 &78.58 & \multicolumn{1}{|c}{54.35} & 56.91 & 55.60 \\
  UMGF-VinVL-EVG \citep{zhang2021multi} & \multicolumn{1}{|c}{51.62} & 51.72 & \multicolumn{1}{c|}{51.67} & 79.02& 78.64& 78.83 & \multicolumn{1}{|c}{55.68} & 55.80 & 55.74 \\
  ITA-VinVL-EVG \citep{wang-etal-2022-ita} & \multicolumn{1}{|c}{52.37} & 50.77 & \multicolumn{1}{c|}{51.56} & 80.40& 78.37& 79.37 & \multicolumn{1}{|c}{56.57} & 54.84 & 55.69 \\
  BARTMNER-VinVL-EVG \citep{yu-etal-2023-grounded} & \multicolumn{1}{|c}{52.47} & 52.43 & \multicolumn{1}{c|}{52.45} & 80.65& 80.14& 80.39 & \multicolumn{1}{|c}{55.68} & 55.63 & 55.66 \\
  H-Index \citep{yu-etal-2023-grounded}  & \multicolumn{1}{|c}{56.16} & 56.67 & \multicolumn{1}{c|}{56.41} & 79.37& 80.10& 79.73 & \multicolumn{1}{|c}{60.90} & 61.46 & 61.18 \\
   \midrule 
    RiVEG$_{\text{BERT}}$ (ours) & \multicolumn{1}{|c}{64.03} & 63.57 & \multicolumn{1}{c|}{63.80} & 83.12& 82.66& 82.89 & \multicolumn{1}{|c}{67.16} & 66.68 & 66.92 \\
  RiVEG$_{\text{XLMR}}$ (ours) & \multicolumn{1}{|c}{65.09} & 66.68 & \multicolumn{1}{c|}{65.88} & 84.80 & 84.23 & 84.51 & \multicolumn{1}{|c}{67.97} & 69.63 & 68.79 \\
  $\bigtriangleup$  & \multicolumn{1}{|c}{\textcolor{red}{$+$8.93}} & \textcolor{red}{$+$10.01} & \multicolumn{1}{c|}{\textcolor{red}{$+$9.47}} & \textcolor{red}{$+$5.43}& \textcolor{red}{$+$4.13}& \textcolor{red}{$+$4.78} & \multicolumn{1}{|c}{\textcolor{red}{$+$7.07}} & \textcolor{red}{$+$8.17} & \textcolor{red}{$+$7.61}\\
   \midrule
   \textbf{RiVEG$_{\text{BERT}}$$^\dag$} (ours) & \multicolumn{1}{|c}{65.10} & 66.96 & \multicolumn{1}{c|}{66.02} & 83.02 & 85.40 & 84.19 & \multicolumn{1}{|c}{68.21} & \textbf{70.18} & 69.18 \\
   
  \textbf{RiVEG$_{\text{XLMR}}$$^\dag$} (ours) & \multicolumn{1}{|c}{\textbf{67.02}} & \textbf{67.10} & \multicolumn{1}{c|}{\textbf{67.06}} & \textbf{85.71}& \textbf{86.16}& \textbf{85.94} & \multicolumn{1}{|c}{\textbf{69.97}} & 70.06 & \textbf{70.01} \\
  $\bigtriangleup^\dag$  & \multicolumn{1}{|c}{\textcolor{red}{$+$10.86}} & \textcolor{red}{$+$10.43} & \multicolumn{1}{c|}{\textcolor{red}{$+$10.65}} & \textcolor{red}{$+$6.34}& \textcolor{red}{$+$6.06}& \textcolor{red}{$+$6.21} & \multicolumn{1}{|c}{\textcolor{red}{$+$9.07}} & \textcolor{red}{$+$8.6} & \textcolor{red}{$+$8.83}\\
  
\bottomrule
\end{tabular}
\caption{Performance comparison on the Twitter-GMNER dataset. All baseline results are from \citet{yu-etal-2023-grounded}. The model marked with $^\dag$ utilizes the data augmentation method. Detailed evaluation metrics are provided in Appendix \ref{evaluation metrics}. And the model configuration details of RiVEG are detailed in Appendix \ref{config}. $\bigtriangleup$ and $\bigtriangleup^\dag$ indicate the performance improvement compared with the previous state-of-the-art method H-Index.}
\label{tab:Main result}
\end{table*}
\paragraph{Visual Grounding Module}
Expansion expressions bring named entities closer to the referring expressions required by existing VG methods. And VE module effectively filters out ungroundable samples. Once there is no requirement for VG methods to assess the groundability of the text input, existing vanilla VG methods can be seamlessly applied to determine the visually grounded region of the named entity. Denote the subset of all groundable named entity samples in GMNER dataset as: 
\vspace{-3pt}
\begin{equation}
    \mathcal{D_{\text{GMNER(VG)}}}=\{(i_1, h_1, v_1), \cdots, (i_k, h_k, v_k)\} \nonumber
    \vspace{-3pt}
\end{equation}
where $i_i$, $h_i$, $v_i$ represent an image that corresponds to a named entity, a named entity referring expression and a visually grounded region of $h_i$, respectively. Note that if a named entity in $\mathcal{D_{\text{GMNER(VG)}}}$ corresponds to multiple different bounding boxes, we specify that the bounding box with the largest area is chosen as the unique gold label. Subsequently, RiVEG employs $\mathcal{D_{\text{GMNER(VG)}}}$ to substitute the original VG dataset and fine-tune the existing vanilla VG model.

\section{Experiments}

\subsection{Settings}
\paragraph{Datasets}
Based on two benchmark MNER datasets, \emph{i.e.}, Twitter-2015 \citep{zhang2018adaptive} and Twitter-2017 \citep{yu-etal-2020-improving-multimodal}, \citet{yu-etal-2023-grounded} builds the Twitter-GMNER dataset by filtering samples with missing images or more than 3 entities belonging to the same type. More details about the training data for different modules and the Twitter-GMNER dataset are provided in Appendix \ref{more dataset}.

\subsection{Main Results}
We compare RiVEG with all existing baseline methods in Table \ref{tab:Main result}. Following \citet{yu-etal-2023-grounded}, we also report results on two subtasks of GMNER, including MNER and Entity Extraction \& Grounding (EEG). MNER aims to identify entity-type pairs, while EEG aims to extract entity-region pairs. The first group of text-only methods focuses on extracting entity-type pairs from text inputs. And the region prediction is then specified as the majority class, \emph{i.e., None}. The second group of multimodal methods comprises the EVG series baseline methods and the end-to-end H-Index method proposed by \citet{yu-etal-2023-grounded}. 

The experimental results demonstrate the superiority of RiVEG over previous methods. Compared with the previous state-of-the-art method H-Index, RiVEG  exhibits significant advantages, achieving absolute improvements of 9.47\%, 4.78\% and 7.61\% in all three tasks without any data augmentation. Additionally, data augmentation methods exhibit significant effects. After data augmentation, RiVEG achieves an additional 1.18\%, 1.43\%, and 1.22\% performance improvement, respectively. This illustrates the rationality and effectiveness of our motivation to maximize the MNER effect and introduce the VE module and entity expansion expressions to use advanced VG methods to solve the GMNER task. We attribute the performance improvement of RiVEG to the following factors: 1) Maximizing the effectiveness of MNER significantly mitigates the issue of error propagation in GMNER predictions. Because if there is a prediction error in the entity or entity type within an entity-type-region prediction triplet, the triplet must be judged as an error. RiVEG alleviates this problem by making the MNER stage independent and introducing auxiliary knowledge. 2) RiVEG does not utilize the OD method for pre-extracting features from candidate regions. As shown in Table \ref{tab:OD test}, existing OD methods exhibit subpar performance on the Twitter-GMNER dataset. RiVEG leverages the feature of VG methods, eliminating the need for pre-extracting features and naturally addressing this deficiency. 3) RiVEG designs an independent VE module to specifically tackle the weak correlation between image-text pairs in GMNER task. This method of enabling the model to undergo targeted training based on data characteristics enables RiVEG to gain a better ability to determine the groundability of named entities. 

Note that this result does not represent the performance upper limit of RiVEG. The modular design facilitates straightforward upgrades, selecting more advanced methods for different modules in the future is expected to yield improved performance.\footnote{Furthermore, we test 10 other text and visual variations of RiVEG in detail in Appendix \ref{visual variants} and \ref{text variants}. Experimental results show that the performance of all 14 RiVEG variants is superior to the previous state-of-the-art method.}

\subsection{Detailed Analysis}

\begin{table}[t!]
\small
\setlength\tabcolsep{0.8pt}
\renewcommand{\arraystretch}{1.2}
\centering
\begin{tabular}{l|cccccc}
\toprule 
\multirow{2}{*}{\textbf{MNER}}
& \multicolumn{3}{c|}{\textbf{Twitter-2015}} & \multicolumn{3}{c}{\textbf{Twitter-2017}}\\
& Pre. & Rec. & \multicolumn{1}{c|}{F1}  & Pre. & Rec. & \multicolumn{1}{c}{F1}    \\
\midrule 
Baseline & 76.45 & 78.22 & \multicolumn{1}{c|}{77.32} & 88.46 & 90.23 & 89.34 \\

ITA(OCR+OD+IC) & - & - & \multicolumn{1}{c|}{78.03} & - & - &  89.75 \\
MoRe$_{\text{Text}}$(Wiki) & - & - & \multicolumn{1}{c|}{77.91} & - & - & 89.50\\
MoRe$_{\text{Image}}$(Wiki) & - & - & \multicolumn{1}{c|}{78.13} & - & - & 89.82 \\
\midrule 
LlaMA2-7B  & 78.16 & 79.57 & \multicolumn{1}{c|}{78.86} & 90.12 & 91.19 & 90.66 \\ 
LlaMA2-13B & 78.37 & \textbf{79.86} & \multicolumn{1}{c|}{79.10} & 89.57 & 92.15 & 90.84 \\
Vicuna-7B  & 76.97 & 79.42 & \multicolumn{1}{c|}{78.17} & 89.35 & 91.27 & 90.30 \\
Vicuna-13B  & 78.46  & 78.92 & \multicolumn{1}{c|}{78.69} & 90.16  & 90.23 & 90.20 \\
PGIM(ChatGPT)    & \textbf{79.21} & 79.45 & \multicolumn{1}{c|}{79.33} & 90.86 & 92.01 & 91.43 \\
\midrule
RiVEG(Ours)  & 79.10  & 79.78 & \multicolumn{1}{c|}{\textbf{79.44}} & \textbf{91.12}  & \textbf{92.67} & \textbf{91.89} \\
 & $\pm$0.46 & $\pm$0.22 & \multicolumn{1}{c|}{$\pm$0.17} & $\pm$0.23 & $\pm$0.09 & $\pm$0.08 \\

\bottomrule
\end{tabular}
\caption{The effectiveness of auxiliary knowledge provided by different LLMs in the MNER stage. The text encoder used by all methods is XLM-RoBERTa$_{\text{large}}$. Baseline means no auxiliary knowledge is added.}
\label{tab:PGIM ChatGPT}
\end{table}

\begin{table}[t!]
\small
\setlength\tabcolsep{2.7pt}
\renewcommand{\arraystretch}{1.2}
\centering
\begin{tabular}{l|cccccc}
\toprule
\multirow{1}{*}{\textbf{LLMs in GMNER$_{\text{VE}}$(Acc.)}} & \multicolumn{1}{c}{\textbf{ALBEF}} & \multicolumn{1}{c}{\textbf{OFA$_\text{large(VE)}$}} \\
\midrule
 Baseline   & 80.39 & 82.08 \\
 Vicuna-7B \citep{chiang2023vicuna} & 82.13 & 82.57 \\
 Vicuna-13B \citep{chiang2023vicuna}& 82.48 & 83.16 \\
 LlaMA2-7B \citep{touvron2023llama} & 82.09 & 83.27 \\
 LlaMA2-13B \citep{touvron2023llama} & 82.62 & 83.75 \\
 ChatGPT \citep{ouyang2022training} & \textbf{83.36} & \textbf{84.30} \\
 Mix & 82.65 & 84.12 \\
\midrule \midrule
\multirow{1}{*}{\textbf{LLMs in GMNER$_{\text{VG}}$(Prec@0.5)}} & \multicolumn{1}{c}{\textbf{SeqTR}} & \multicolumn{1}{c}{\textbf{OFA$_\text{large(VG)}$}} \\ \midrule
  Baseline  & 67.27 & 72.40 \\
  Vicuna-7B \citep{chiang2023vicuna} & 71.43 & 73.05 \\
 Vicuna-13B \citep{chiang2023vicuna} & 71.51 & 73.26 \\
 LlaMA2-7B \citep{touvron2023llama} & 70.98 & 72.89 \\
 LlaMA2-13B \citep{touvron2023llama}& 71.39 & 73.45 \\
 ChatGPT \citep{ouyang2022training} & 72.10 & 73.90 \\
 Mix & \textbf{73.06} & \textbf{74.50} \\
\bottomrule
\end{tabular}
\caption{Results of the same model on datasets built by different LLMs. Baseline uses original named entities and their entity types. The rest of tests use their respective generated named entity referring expression.}
\label{LLM in VE VG}
\end{table}

\paragraph{Contributions of Different LLMs to Various Subtasks}
Table \ref{tab:PGIM ChatGPT} shows the test set results after using the same MNER model trained on different LLM versions of the same dataset. The method of obtaining external knowledge is exactly the same as PGIM \citep{li-etal-2023-prompting}. Compared with ITA \citep{wang-etal-2022-ita} (Using optical character recognition, object detection and image caption) and MoRe \citep{wang-etal-2022-named} (Using Wikipedia), the auxiliary knowledge provided by LLMs is more helpful to MNER. The quality of knowledge provided by LLMs with a small number of parameters is usually inferior to ChatGPT used by PGIM. RiVEG performs better than past state-of-the-art methods.\footnote{This is average result from three different random seeds.} The training set of RiVEG includes responses from all five LLMs. And the dev set and test set are the same as PGIM. This indicates that utilizing diverse styles of external knowledge based on the same original sample enhances the generalization ability of model. 

The named entities and their visual grounding results in the Twitter-GMNER dataset are deterministic, so using different LLMs to generate different named entity expansion expressions can build different LLM versions of the dataset. Table \ref{LLM in VE VG} shows the test set results of the same VE and VG models after training on different versions of the datasets. The training set of the Mix version is a direct merger of all five training sets, and the dev set is identical to the ChatGPT version. Using all versions of the named entity referring expression as text input performs significantly better than using raw named entities. LLMs with a large number of parameters generally generate better quality results than LLMs with a small number of parameters. The results of the Mix version show that more different styles of generated data can improve model performance. This is reflected more intuitively in Appendix \ref{text variants}.

\paragraph{Named Entity Referring Expressions Analysis}

\begin{table}[t!]
\small
\setlength\tabcolsep{2.1pt}
\renewcommand{\arraystretch}{1.2}
\centering
\begin{tabular}{l|cccccc}
\toprule
\multirow{1}{*}{\textbf{VE Module Text Input (Acc.)}} & \multicolumn{1}{c}{\textbf{ALBEF}} & \multicolumn{1}{c}{\textbf{OFA$_\text{large(VE)}$}} \\
\midrule
 Named Entity+Entity Category& 80.39 & 82.08 \\
 Entity Expansion Expression& 81.76 & 81.40 \\
 Named Entity Referring Expression& \textbf{83.36} & \textbf{84.30}  \\\midrule \midrule
\multirow{1}{*}{\textbf{VG Module Text Input (Prec@0.5)}} & \multicolumn{1}{c}{\textbf{SeqTR}} & \multicolumn{1}{c}{\textbf{OFA$_\text{large(VG)}$}} \\
\midrule
 Named Entity+Entity Category& 67.27 & 72.40 \\
 Entity Expansion Expression& 71.72 & 71.10  \\
 Named Entity Referring Expression& \textbf{72.10} & \textbf{73.90}  \\
\bottomrule

\end{tabular}
\caption{The influence of varying text inputs on the performance of different methods.}
\label{tab:text analysis}
\end{table}

We further investigate the performance of various methods in VE and VG modules when confronted with diverse text inputs. The named entity referring expression in $\mathcal{D_{\text{GMNER(VE/VG)}}}$ is segmented into the named entity along with entity category and the entity expansion expression. Table \ref{tab:text analysis} shows the test set performance of different methods after fine-tuning on $\mathcal{D_{\text{GMNER(VE/VG)}}}$. In order to control variables, this experiment is conducted on the expansion expression generated by ChatGPT. 

The experimental results demonstrate that when the text inputs are the complete named entity referring expressions, all methods achieve optimal results. When the text inputs are named entities or entity expansion expressions, the performance of all methods decreases to a certain extent. Moreover, the sensitivity of different methods to various text inputs varies significantly. Unified frameworks like OFA demonstrate significantly better performance on fine-grained named entities than on coarse-grained entity expansion expressions. Contrarily, specialized methods like SeqTR, designed for a specific task, exhibit the opposite behavior. This experiment verifies the rationality of our proposed named entity referring expression. While different methods exhibit distinct characteristics, the expression method designed by us that combines both coarse and fine granularity emerges as their optimal choice.

\paragraph{Effect Analysis of Additional Training Data}

\begin{table}[t!]
\footnotesize
\setlength\tabcolsep{0.4pt}
\renewcommand{\arraystretch}{1.2}
\centering
\begin{tabular}{l|cccccc}
\toprule
\multirow{1}{*}{\textbf{VE Training Data}} & \multicolumn{1}{c}{OFA$_\text{base(VE)}$} &\multicolumn{1}{c}{OFA$_\text{large(VE)}$} \\
\midrule
Pretraining+SNLI-VE+$\mathcal{D_{\text{GMNER(VE)}}}$     & 82.90& 83.32 \\
Pretraining+$\mathcal{D_{\text{GMNER(VE)}}}$             & \textbf{83.41}& \textbf{84.30} \\
\midrule \midrule
\multirow{1}{*}{\textbf{VG Training Data}} & \multicolumn{1}{c}{OFA$_\text{base(VG)}$} &\multicolumn{1}{c}{OFA$_\text{large(VG)}$} \\
\midrule
Pretraining+RefCOCOg+$\mathcal{D_{\text{GMNER(VG)}}}$    & 71.60& 73.64 \\
Pretraining+RefCOCO++$\mathcal{D_{\text{GMNER(VG)}}}$   & 71.91& 73.50 \\
Pretraining+RefCOCO+$\mathcal{D_{\text{GMNER(VG)}}}$   & \textbf{72.40}& 72.52 \\
 Pretraining+$\mathcal{D_{\text{GMNER(VG)}}}$            & 68.83& \textbf{73.90} \\
\bottomrule
\end{tabular}
\caption{Performance after fine-tuning with various versions of pretrained weights on different datasets.}
\label{tab:more data}
\end{table}

As the $\mathcal{D_{\text{GMNER(VE/VG)}}}$ datasets we create differ only in text type from the SNLI-VE \citep{xie2019visual} and RefCOCO series datasets \citep{mao2016generation, yu2016modeling, nagaraja2016modeling}, we investigate whether pretraining with more similar task data could yield additional performance improvements in Table \ref{tab:more data}. 

For various VE methods, incorporating additional fine-tuning on the SNLI-VE dataset \citep{xie2019visual} prior to the final fine-tuning on the $\mathcal{D_{\text{GMNER(VE)}}}$ does not result in improved performance. 
This observation suggests a significant difference between the $\mathcal{D_{\text{GMNER(VE)}}}$ and the SNLI-VE. 
For various VG methods, conducting additional fine-tuning on the RefCOCO series of datasets \citep{mao2016generation, yu2016modeling, nagaraja2016modeling} prior to the final fine-tuning on the $\mathcal{D_{\text{GMNER(VG)}}}$ may result in enhanced performance. This may be attributed to a certain degree of similarity between the two. Because we convert named entities into named entity referring expressions through entity expansion expressions.\footnote{ALBEF \citep{li2021align} authors do not provide pretrained weights based on the SNLI-VE dataset. And some of the pretrained weights of SeqTR \citep{zhu2022seqtr} based on the RefCOCO series dataset cannot be obtained. Therefore, we follow the wishes of author and only test OFA\citep{wang2022ofa} in this experiment.}

\section{Conclusion}
In this paper, we propose RiVEG, a unified framework that aims to use LLMs as bridges to reformulate the GMNER task into a unified MNER-VE-VG task. This reformulation enables the framework to seamlessly incorporate the capabilities of state-of-the-art methods for each subtask in their corresponding modules. Extensive experiments show that RiVEG significantly outperforms state-of-the-art methods. The proposed RiVEG greatly reduces the constraints of VG and VE methods on natural language inputs, expands their applicability to more realistic scenarios, and provides insights for future work to better deploy VG and VE modules in related tasks.

\section*{Limitations}
In the past, the three-stage reformulation of GMNER was not promising by researchers. Because there may be two intuitive limitations to this reformulation. Limitation 1. Splitting the entire pipeline into three stages can lead to severe error propagation. Limitation 2. Pipeline design seems to be quite heavy. But actually, regarding Limitation 1, this paper demonstrates for the first time that the three-stage pipeline design performs significantly better than existing end-to-end methods. Regarding Limitation 2, the modular nature of RiVEG allows it to freely select lighter sub-models to form more efficient variants to deal with time-critical scenarios. The most lightweight variant of RiVEG can complete the entire three-stage training in 5 hours using a single 4090 GPU, making the training cost quite low. And the inference speed of RiVEG is also acceptable, since LLMs can be called in the form of API. Considering the substantial performance improvement, sacrificing a certain degree of inference speed is worthwhile. In the future, our research direction is to strive for optimal performance while designing each module in a lightweight style and training them in conjunction, possibly incorporating intermediate auxiliary losses.

\section*{Ethics Statement}
In this paper, we use publicly available Twitter-GMNER dataset for experiments. For the auxiliary refined knowledge and the entity expansion expression, RiVEG generates them using ChatGPT, Vicuna, LlaMA2. Therefore, we trust that all the data we use does not violate the privacy of any user.

\section*{Acknowledgments}
The authors would like to thank four anonymous reviewers for their insightful comments. This work was supported by the Natural Science Foundation of Tianjin (No.21JCYBJC00640) and the 2023 CCF-Baidu SongGuo Foundation (Research on Scene Text Recognition Based on PaddlePaddle).

\bibliography{cite}

\appendix
\section{Appendix}

\subsection{Model Configuration}
\label{config}
In order to reasonably evaluate the capabilities of RiVEG, we choose advanced methods for each of three sub-modules of RiVEG. Specifically, for MNER module, RiVEG uses the same architecture as PGIM \citep{li-etal-2023-prompting}, and selects BERT$_\text{base}$ \citep{devlin-etal-2019-bert} and XLM-RoBERTa$_\text{large}$ \citep{conneau-etal-2020-unsupervised} as the text encoder, aligning with the prevailing MNER methods \citep{wang-etal-2022-named, wang2022promptmner, wang2022cat}. 
mPLUG-Owl \citep{ye2023mplug} is employed to generate the image caption necessary for the entity expansion expression. 
For VE module, on the basis of OFA$_\text{large}$ \citep{wang2022ofa}, RiVEG replaces its original VE dataset SNLI-VE \citep{xie2019visual} with $\mathcal{D_{\text{GMNER(VE)}}}$ and $\mathcal{D_{\text{GMNER(VE)}^\dag}}$ for fine-tuning. 
For VG module, RiVEG chooses OFA$_\text{large}$ \citep{wang2022ofa}, which also possesses visual grounding capability, to substitute its original VG datasets RefCOCO \citep{yu2016modeling}, RefCOCO+ \citep{yu2016modeling}, and RefCOCOg \citep{mao2016generation, nagaraja2016modeling} with $\mathcal{D_{\text{GMNER(VG)}}}$ and $\mathcal{D_{\text{GMNER(VG)}^\dag}}$ for fine-tuning.

\subsection{Details of Different Module Datasets}
\label{more dataset}

\begin{table*}[!]
\small
\setlength\tabcolsep{5.2pt}
\renewcommand{\arraystretch}{1.1}
\centering
\begin{tabular}{l|cccccc}
\toprule 
\multirow{2}{*}{\textbf{Split}} 
 &
\multicolumn{2}{c|}{\textbf{MNER Module Datasets}}  & \multicolumn{2}{c|}{\textbf{VE Module Datasets}} & \multicolumn{2}{c}{\textbf{VG Module Datasets}}  \\
\cline{2-7}
 & \multicolumn{1}{c}{\textbf{\#Twitter-GMNER}} & \multicolumn{1}{c|}{\textbf{\#Twitter-GMNER}$^\dag$} & {\textbf{\#$\mathcal{D_{\text{GMNER(VE)}}}$}} & \multicolumn{1}{c|}{\textbf{\#$\mathcal{D_{\text{GMNER(VE)}^\dag}}$}} & {\textbf{\#$\mathcal{D_{\text{GMNER(VG)}}}$}} & \multicolumn{1}{c}{\textbf{\#$\mathcal{D_{\text{GMNER(VG)}^\dag}}$}} \\ 
  \midrule 
Train & 7000 & \multicolumn{1}{c|}{35000} & 11782 & \multicolumn{1}{c|}{58910} & 4694 & 23470\\
Dev   & 1500 & \multicolumn{1}{c|}{1500}  & 2453  & \multicolumn{1}{c|}{2453} & 986 & 986\\
Test  & 1500 & \multicolumn{1}{c|}{1500}  & 2543 & \multicolumn{1}{c|}{2543} & 1036 & 1036 \\
\midrule
Total & 10000  &  \multicolumn{1}{c|}{38000} & 16778 &  \multicolumn{1}{c|}{63906} & 6716 & 25492  \\
\bottomrule
\end{tabular}
\caption{Statistics of datasets used in training of different sub-modules.}
\label{Different Module Datasets}
\end{table*}

\begin{table}[t!]
\small
\setlength\tabcolsep{4.9pt}
\renewcommand{\arraystretch}{1.0}
\centering
\begin{tabular}{l|cccc}
\toprule
\textbf{Split} & \textbf{\#Tweet}& \textbf{\#Entity} & \textbf{\#Groundable Entity}  & \textbf{\#Box} \\
\midrule
Train & 7000 & 11782 & 4694 & 5680 \\
Dev   & 1500 & 2453  & 986  & 1166 \\
Test  & 1500 & 2543  & 1036 & 1244  \\
\midrule
Total & 10000  & 16778 & 6716 & 8090   \\
\bottomrule
\end{tabular}
\caption{Statistics of Twitter-GMNER dataset \citep{yu-etal-2023-grounded}.}
\label{tab:GMNER dataset}
\end{table}

Table \ref{Different Module Datasets} counts the training data used by different modules. And Table \ref{tab:GMNER dataset} shows the details of the Twitter-GMNER dataset. For the MNER module of RiVEG, the training data for base version comprises original samples and corresponding auxiliary refined knowledge generated by $\text{gpt-3.5-turbo}$ provided by PGIM \citep{li-etal-2023-prompting}. Additionally, data augmentation is applied to the training set of base version. In data augmentation version, we apply four additional LLMs (\emph{i.e.}, \text{vicuna-7b-v1.5}, \text{vicuna-13b-v1.5}, \text{llama-2-7b-chat-hf}, \text{llama-2-13b-chat-hf}) to training set samples using the same algorithm, generating multiple versions of auxiliary refined knowledge for the same sample. The dev set and test set remain the same as base version to ensure the benchmarking of evaluation. 

For VE module, the training data of base version $\mathcal{D_{\text{GMNER(VE)}}}$ exclusively comprises entity expansion expressions generated by $\text{gpt-3.5-turbo}$. In the data augmentation version $\mathcal{D_{\text{GMNER(VE)}^\dag}}$, we also apply four additional LLMs to generate additional training samples and maintain the dev set and test set identical to base version $\mathcal{D_{\text{GMNER(VE)}}}$. 

For VG module, the training data of base version $\mathcal{D_{\text{GMNER(VG)}}}$ is a subset of all groundable entity expansion expressions in base version $\mathcal{D_{\text{GMNER(VE)}}}$. Similarly, the data augmentation version $\mathcal{D_{\text{GMNER(VG)}^\dag}}$ is a subset of the data augmentation version $\mathcal{D_{\text{GMNER(VE)}^\dag}}$.

\subsection{Detailed Implementation Details}
For the methods used by three modules of RiVEG, we only replace their original VE and VG datasets with $\mathcal{D_{\text{GMNER(VE)}}}$/$\mathcal{D_{\text{GMNER(VE)}^\dag}}$ and $\mathcal{D_{\text{GMNER(VG)}}}$/$\mathcal{D_{\text{GMNER(VG)}^\dag}}$ and primarily adhere to original hyperparameter configurations and training strategies as reported in their paper for fine-tuning. Without specific instructions, all training is conducted using a single 4090 GPU. The model with the best results on each dev set is selected to evaluate the performance on each test set and used to perform the final inference. Note that considering the notable effectiveness of RiVEG, we refrain from conducting attempts with excessive hyperparameter combinations. Thus, exploring more hyperparameter combinations may yield improved results.

\paragraph{MNER Module}
RiVEG mainly follows MNER fine-tuning strategy of PGIM\footnote{\url{https://github.com/JinYuanLi0012/PGIM}} \citep{li-etal-2023-prompting}. For the XLM-RoBERTa$_\text{large}$ version in Table \ref{tab:Main result}, Table \ref{tab:other variants} and Table \ref{tab:other text encoder}, learning rate is set to 7e-6 and batchsize is set to 4. For the BERT$_\text{base}$ version in Table \ref{tab:other text encoder}, learning rate is set to 5e-5 and batchsize is set to 16. Remaining settings remain unchanged. The model is fine-tuned for 25 epochs. 

\paragraph{VE Module}
For OFA$_\text{large(VE)}$ \citep{wang2022ofa}, the learning rate is set to 2e-5 and batchsize is set to 32. We convert the labels entailment/contradiction to yes/no. And we also use the Trie-based search strategy to constrain the generated labels over the candidate set.  Remaining settings follow their VE fine-tuning strategy\footnote{\url{https://github.com/OFA-Sys/OFA\#visual-entailment}}. The basic version of OFA$_\text{large}$ is fine-tuned for 6 epochs on $\mathcal{D_{\text{GMNER(VE)}}}$/$\mathcal{D_{\text{GMNER(VE)}^\dag}}$. 

For ALBEF \citep{li2021align}, the learning rate is set to 2e-5 and batchsize is set to 24. Remaining settings follow their VE fine-tuning strategy\footnote{\url{https://github.com/salesforce/ALBEF\#visual-entailment}}. We fine-tune for 5 epochs on $\mathcal{D_{\text{GMNER(VE)}}}$/$\mathcal{D_{\text{GMNER(VE)}^\dag}}$ using a single V100 32G GPU based on ALBEF-14M. 

\begin{table*}[!]
\small
\setlength\tabcolsep{3.6pt}
\renewcommand{\arraystretch}{1.1}
\centering
\begin{tabular}{lcccccccccc}
\toprule
\multicolumn{1}{c}{\textbf{Visual Variants of RiVEG}} 
 & \multicolumn{3}{|c|}{\textbf{GMNER}}  & \multicolumn{3}{c|}{\textbf{MNER}} & \multicolumn{3}{c}{\textbf{EEG}}  \\
\cline{2-10}
\multicolumn{1}{c}{MNER(F1)+VE(Acc.)+VG(Prec@0.5)}
 & \multicolumn{1}{|c}{\textbf{Pre.}} & \textbf{Rec.} & \multicolumn{1}{c|}{\textbf{F1}} & \textbf{Pre.} & \textbf{Rec.} & \multicolumn{1}{c|}{\textbf{F1}} & \textbf{Pre.} & \textbf{Rec.} & \multicolumn{1}{c}{\textbf{F1}} \\ 
  \midrule 
  Baseline (H-Index \citep{yu-etal-2023-grounded}) & \multicolumn{1}{|c}{56.16} & 56.67 & \multicolumn{1}{c|}{56.41} & 79.37 & 80.10& 79.73 & \multicolumn{1}{|c}{60.90} & 61.46 & 61.18
 \\
 \midrule 
 \multicolumn{10}{c}{ \emph{w/o Data Augmentation}}\\
 \midrule
 PGIM(84.51)+OFA$_\text{large(VE)}$(84.30)+SeqTR(72.10) & \multicolumn{1}{|c}{57.06} & 58.46 & \multicolumn{1}{c|}{57.75} & 84.80 & 84.23 & 84.51 & \multicolumn{1}{|c}{59.79} & 61.25 & 60.51 \\ 
 PGIM(84.51)+ALBEF-14M(83.36)+SeqTR(72.10) & \multicolumn{1}{|c}{57.45} & 58.85 & \multicolumn{1}{c|}{58.14} & 84.80 & 84.23 & 84.51 & \multicolumn{1}{|c}{60.25} & 61.72 & 60.98 \\
 PGIM(84.51)+ALBEF-14M(83.36)+OFA$_\text{large(VG)}$(73.90) & \multicolumn{1}{|c}{64.56} & 66.13 & \multicolumn{1}{c|}{65.33} & 84.80 & 84.23 & 84.51 & \multicolumn{1}{|c}{67.47} & 69.12 & 68.29 \\
 PGIM(84.51)+OFA$_\text{large(VE)}$(84.30)+OFA$_\text{large(VG)}$(73.90) & \multicolumn{1}{|c}{65.09} & 66.68 & \multicolumn{1}{c|}{65.88} & 84.80 & 84.23 & 84.51 & \multicolumn{1}{|c}{67.97} & 69.63 & 68.79 \\
 \midrule 
 \multicolumn{10}{c}{ \emph{ Data Augmentation}}\\
 \midrule
 PGIM(85.94)+OFA$_\text{large(VE)}$(84.12)+SeqTR(73.06) & \multicolumn{1}{|c}{59.03} & 59.10 & \multicolumn{1}{c|}{59.06} & 85.71 & 86.16 & 85.94 & \multicolumn{1}{|c}{61.85} & 61.93 & 61.89 \\ 
 PGIM(85.94)+ALBEF-14M(82.65)+SeqTR(73.06) & \multicolumn{1}{|c}{59.14} & 59.22 & \multicolumn{1}{c|}{59.18} & 85.71 & 86.16 & 85.94 & \multicolumn{1}{|c}{62.01} & 62.09 & 62.05 \\ 
 PGIM(85.94)+ALBEF-14M(82.65)+OFA$_\text{large(VG)}$(74.50) & \multicolumn{1}{|c}{66.12} & 66.20 & \multicolumn{1}{c|}{66.16} & 85.71 & 86.16 & 85.94 & \multicolumn{1}{|c}{68.99} & 69.07 & 69.03 \\ 
 PGIM(85.94)+OFA$_\text{large(VE)}$(84.12)+OFA$_\text{large(VG)}$(74.50) & \multicolumn{1}{|c}{67.02} & 67.10 & \multicolumn{1}{c|}{67.06} & 85.71 & 86.16 & 85.94 & \multicolumn{1}{|c}{69.97} & 70.06 & 70.01 \\ 
 
\bottomrule
\end{tabular}
\caption{Performance comparison of different visual variants of RiVEG. Here we keep MNER module unchanged to clearly illustrate the impact of different VE and VG methods on performance. Therefore, different variants of the same group have the same MNER results. The test results of more text variants are shown in Table \ref{tab:other text encoder}.}
\label{tab:other variants}
\end{table*}

\paragraph{VG Module}
For OFA$_\text{large(VG)}$ \citep{wang2022ofa}, learning rate is 3e-5 with the label smoothing of 0.1 and batchsize is set to 32. The input image resolution is set to 512$\times$512. And the basic version of OFA$_\text{large}$ is fine-tuned for 10 epochs on $\mathcal{D_{\text{GMNER(VG)}}}$/$\mathcal{D_{\text{GMNER(VG)}^\dag}}$. Remaining settings follow their VG fine-tuning strategy\footnote{\url{https://github.com/OFA-Sys/OFA\#visual-grounding-referring-expression-comprehension}}.

For SeqTR \citep{zhu2022seqtr}, since SeqTR does not provide pretrained weights, we select the weight which are pretrained and fine-tuned for 5 epochs on RefCOCO\footnote{\url{https://github.com/seanzhuh/SeqTR\#refcoco}} \citep{yu2016modeling}, as the initial weight and fine-tune for 5 epochs on $\mathcal{D_{\text{GMNER(VG)}}}$/$\mathcal{D_{\text{GMNER(VG)}^\dag}}$. The input image resolution is set to 640$\times$640. Text length is trimmed to 30. The learning rate is 5e-4 and batchsize is 64. Visual encoder is YOLOv3 \citep{redmon2018yolov3}. Remaining settings follow their Referring Expression Comprehension fine-tuning strategy\footnote{\url{https://github.com/seanzhuh/SeqTR\#pre-training--fine-tuning}}.

\subsection{Evaluation Metrics}
\label{evaluation metrics}
The GMNER prediction is composed of entity, type, and visual region. Following \citet{yu-etal-2023-grounded}, the correctness of each prediction is computed as follows: 

\vspace{-3pt}
\begin{align}
    C_e/C_t &= 
    \begin{cases}
        1, & p_e/p_t = g_e/g_t; \\
        0, & \text{otherwise.}
    \end{cases}
    \nonumber \\
    C_v &= 
    \begin{cases}
        1, & p_v = g_v = None; \\
        1, & \text{max(}IoU_1,\cdots,IoU_j\text{)} > 0.5; \\
        0, & \text{otherwise.}  
    \end{cases}
    \nonumber \\
    correct &= 
    \begin{cases}
        1, & C_e\land C_t\land C_v = 1; \\
        0, & \text{otherwise.}
    \end{cases}
    \nonumber
    \vspace{-3pt}
\end{align}
where $C_e$, $C_t$ and $C_v$ represent the correctness of entity, type and region predictions; $p_e$, $p_t$ and $p_v$ represent the predicted entity, type and region; $g_e$, $g_t$ and $g_v$ represent the gold entity, type and region; and $IoU_j$ denotes the IoU score between $p_v$ with the $j$-th ground-truth bounding box $g_{v,j}$. Then precision (Pre.), recall (Rec.) and F1 score are used to evaluate the performance of model: 
\vspace{-3pt}
\begin{align}
     Pre = \frac{\#correct}{\#predict}, Rec = \frac{\#correct}{\#gold}\nonumber
    \vspace{-8pt}
\end{align}
\vspace{-8pt}
\begin{align}
    F1 = \frac{2\times Pre\times Rec}{Pre+Rec}
    \nonumber
    \vspace{-3pt}
\end{align}
where $\#correct$, $\#predict$ and $\#gold$ respectively represent the number of triples of correct predictions, predictions and gold labels.

\subsection{Exploration of More Visual Variants}

\begin{table*}[!]
\small
\setlength\tabcolsep{2.0pt}
\renewcommand{\arraystretch}{1.1}
\centering
\begin{tabular}{lcccccccccc}
\toprule
\multicolumn{1}{c}{\textbf{Text Variants of RiVEG}} 
 & \multicolumn{3}{|c|}{\textbf{GMNER}}  & \multicolumn{3}{c|}{\textbf{MNER}} & \multicolumn{3}{c}{\textbf{EEG}}  \\
\cline{2-10}
\multicolumn{1}{c}{MNER(F1)+VE(Acc.)+VG(Prec@0.5)}
 & \multicolumn{1}{|c}{\textbf{Pre.}} & \textbf{Rec.} & \multicolumn{1}{c|}{\textbf{F1}} & \textbf{Pre.} & \textbf{Rec.} & \multicolumn{1}{c|}{\textbf{F1}} & \textbf{Pre.} & \textbf{Rec.} & \multicolumn{1}{c}{\textbf{F1}} \\ 
 
 \midrule 
 \multicolumn{10}{c}{ \emph{ Baseline}}\\
 \midrule 
  Unimodal Baseline (BERT-CRF-None \citep{yu-etal-2023-grounded})$^\ddag$ & \multicolumn{1}{|c}{42.73} & 44.88 & \multicolumn{1}{c|}{43.78} & 77.23 & 78.64 & 77.93 & \multicolumn{1}{|c}{46.92} & 49.28 & 48.07 \\
  Multimodal Baseline (H-Index \citep{yu-etal-2023-grounded})$^\ddag$ & \multicolumn{1}{|c}{56.16} & 56.67 & \multicolumn{1}{c|}{56.41} & 79.37 & 80.10& 79.73 & \multicolumn{1}{|c}{60.90} & 61.46 & 61.18 
 \\  
 
 \midrule 
 \multicolumn{10}{c}{ \emph{w/o Data Augmentation}}\\
 \midrule
 PGIM$_\text{BERT}$(82.89)+OFA$_\text{large(VE)}$(84.30)+OFA$_\text{large(VG)}$(73.90) & \multicolumn{1}{|c}{64.03} & 63.57 & \multicolumn{1}{c|}{63.80} & 83.12 & 82.66 & 82.89 & \multicolumn{1}{|c}{67.16} & 66.68 & 66.92 \\
 PGIM$_\text{XLMR}$(84.51)+OFA$_\text{large(VE)}$(84.30)+OFA$_\text{large(VG)}$(73.90) & \multicolumn{1}{|c}{65.09} & 66.68 & \multicolumn{1}{c|}{65.88} & 84.80 & 84.23 & 84.51 & \multicolumn{1}{|c}{67.97} & 69.63 & 68.79 \\
 
 \midrule 
 \multicolumn{10}{c}{ \emph{ Data Augmentation}}\\
 \midrule 
 PGIM$_\text{BERT}$(84.19)+OFA$_\text{large(VE)}$(84.12)+OFA$_\text{large(VG)}$(74.50) & \multicolumn{1}{|c}{65.10} & 66.96 & \multicolumn{1}{c|}{66.02} & 83.02 & 85.40 & 84.19 & \multicolumn{1}{|c}{68.21} & \textbf{70.18} & 69.18 \\
 \textbf{PGIM$_\text{XLMR}$(85.94)+OFA$_\text{large(VE)}$(84.12)+OFA$_\text{large(VG)}$(74.50)} &  \multicolumn{1}{|c}{\textbf{67.02}} & \textbf{67.10} & \multicolumn{1}{c|}{\textbf{67.06}} & \textbf{85.71} & \textbf{86.16} & \textbf{85.94} & \multicolumn{1}{|c}{\textbf{69.97}} & 70.06 & \textbf{70.01} \\ 
\midrule 
 \multicolumn{10}{c}{ \emph{w/o Auxiliary Refined Knowledge in MNER}}\\
 \midrule 
   BERT-CRF(78.19)+OFA$_\text{large(VE)}$(84.30)+OFA$_\text{large(VG)}$(73.90)  & \multicolumn{1}{|c}{61.16} & 59.95 & \multicolumn{1}{c|}{60.55} & 78.93 & 77.47 & 78.19 & \multicolumn{1}{|c}{65.33} & 64.04 & 64.68
 \\
    BERT-CRF(78.19)+OFA$_\text{large(VE)}$(84.12)$^\dag$+OFA$_\text{large(VG)}$(74.50)$^\dag$  & \multicolumn{1}{|c}{62.16} & 60.94 & \multicolumn{1}{c|}{61.54} & 78.93 & 77.47 & 78.19 & \multicolumn{1}{|c}{66.29} & 64.99 & 65.63
 \\     
    XLMR-CRF(82.90)+OFA$_\text{large(VE)}$(84.30)+OFA$_\text{large(VG)}$(73.90)  & \multicolumn{1}{|c}{63.89} & 64.24 & \multicolumn{1}{c|}{64.06} & 82.68 & 83.13 & 82.90 & \multicolumn{1}{|c}{67.45} & 67.82 & 67.63
 \\
     XLMR-CRF(82.90)+OFA$_\text{large(VE)}$(84.12)$^\dag$+OFA$_\text{large(VG)}$(74.50)$^\dag$  & \multicolumn{1}{|c}{64.67} & 65.03 & \multicolumn{1}{c|}{64.85} & 82.68 & 83.13 & 82.90 & \multicolumn{1}{|c}{68.11} & 68.49 & 68.30
 \\
 
\bottomrule
\end{tabular}
\caption{Performance of different text variants of RiVEG. Different from Table \ref{tab:other variants}, here we keep the VE and VG modules unchanged. For the baseline model, results of methods with $^\ddag$ come from \citet{yu-etal-2023-grounded}. $^\dag$ indicates that the VE and VG modules use data augmentation methods.}
\label{tab:other text encoder}
\end{table*}

To further demonstrate the effectiveness of RiVEG, we present the performance of various RiVEG variants across all three subtasks in Table \ref{tab:other variants}. Specifically, we additionally choose an early classic multimodal pretraining method ALBEF-14M \citep{li2021align} for VE module and an early classic VG method SeqTR \citep{zhu2022seqtr} for VG module. Different combinations of methods are considered as different variants. Experimental results show that, without considering any data augmentation, the weakest variant still exhibits highly competitive results with the previous state-of-the-art method. And all the variants after data augmentation are significantly better than the baseline. 

We also present the corresponding test set results for different methods of different modules after fine-tuning. The results indicate that stronger combinations of sub-methods generally yield superior performance. The performance of the early classic ALBEF and SeqTR is inferior to the subsequent advanced OFA. Additionally, while the test set Prec@0.5 of SeqTR is slightly lower than that of OFA$_\text{large(VG)}$, there are significant differences in their final GMNER and EEG results. This phenomenon may be attributed to the text representation method employed by SeqTR. 
The GRU \citep{chung2014empirical} vocabulary of SeqTR is solely statistically obtained from $\mathcal{D_{\text{GMNER(VG)}}}$ or $\mathcal{D_{\text{GMNER(VG)}^\dag}}$, rendering it incapable of handling out-of-vocabulary words in new samples during inference. 

Moreover, data augmentation methods exhibit varying effects on distinct modules. MNER and VG methods can intuitively benefit from data augmentation, but this is not the case for VE methods. One potential reason is that the distribution after fitting more training data does not align with the optimal distribution of the test data. But this is acceptable since data augmentation significantly enhances the performance of overall framework. 
\label{visual variants}

\subsection{Exploration of More Text Variants}
\label{text variants}
Table \ref{tab:other text encoder} explores the influence of various text encoders on the framework. Specifically, we fix the VE and VG modules and replace the originally used XLM-RoBERTa$_\text{large}$ in the MNER module with BERT$_\text{base}$. The main conclusions are as follows: 1) The weaker text encoder evidently results in inferior MNER performance compared with the XLM-RoBERTa variant. Due to the error propagation characteristics of GMNER prediction triples, weak MNER performance naturally results in the degradation of GMNER and EEG performance. 2) Concerning the MNER results, the BERT version of RiVEG outperforms the BERT version of Unimodal Baseline, surpassing 4.96\% without any data augmentation and achieving a further lead of 6.26\% after data augmentation. It underscores the effectiveness of incorporating auxiliary knowledge into the original samples in MNER task. 

Additionally, to control variables and compare entity grounding performance across different methods, we conduct more challenging tests for RiVEG. The variants in the \emph{w/o Auxiliary Refined Knowledge in MNER} group indicate that the text input to their MNER module consists only of the original text and does not incorporate any external knowledge. In this scenario, the MNER performance of these BERT variants of RiVEG is nearly identical to that of the two baseline methods. However, the GMNER and EEG performance of these variants still significantly outperforms all baseline methods. This comparison highlights the robust performance of RiVEG in entity grounding. 

The last four lines of experimental results also illustrate the efficacy of the data augmentation method. 
When facing the same named entity recognition and expansion results, VE and VG models trained using data augmentation methods generally achieve better GMNER and EEG results. This echoes the experimental results in Table \ref{LLM in VE VG}.

\subsection{Case Study}
\begin{figure*}
	\centering
	\includegraphics[scale=0.475]{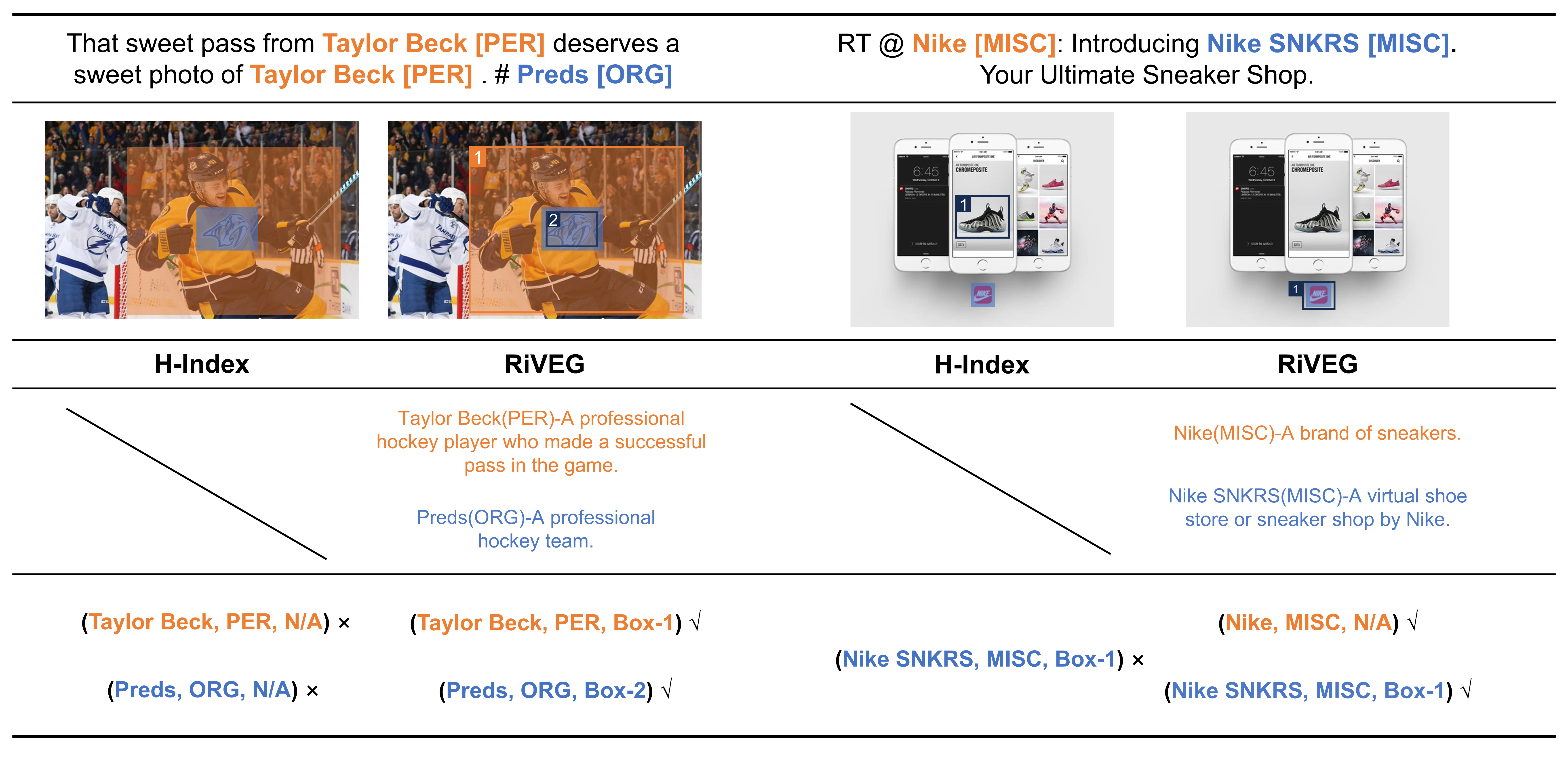}
	\caption{Compare the predictions of RiVEG and H-Index for two test samples.}
	\label{fig:cases right}
\end{figure*}

We further conduct case study to compare the predictions of H-Index \citep{yu-etal-2023-grounded} and RiVEG for challenging samples. Figure \ref{fig:cases right} illustrates two examples where H-Index makes errors, while RiVEG produces accurate predictions. 
For the test sample on the left, even though H-Index correctly predicts the entity-type pairs, it fails to achieve any effective entity grounding. All predictions made by RiVEG are accurate, even for the challenging entity grounding case of the ``\emph{Preds}'' team logo. 
For the sample on the right, H-Index fails to predict all named entities and does not perform entity grounding correctly. RiVEG completes predictions perfectly, even though the grounded area of ``$\emph{Nike\ \allowbreak SNKRS}$'' is very small. 

\begin{figure*}
	\centering
	\includegraphics[scale=0.475]{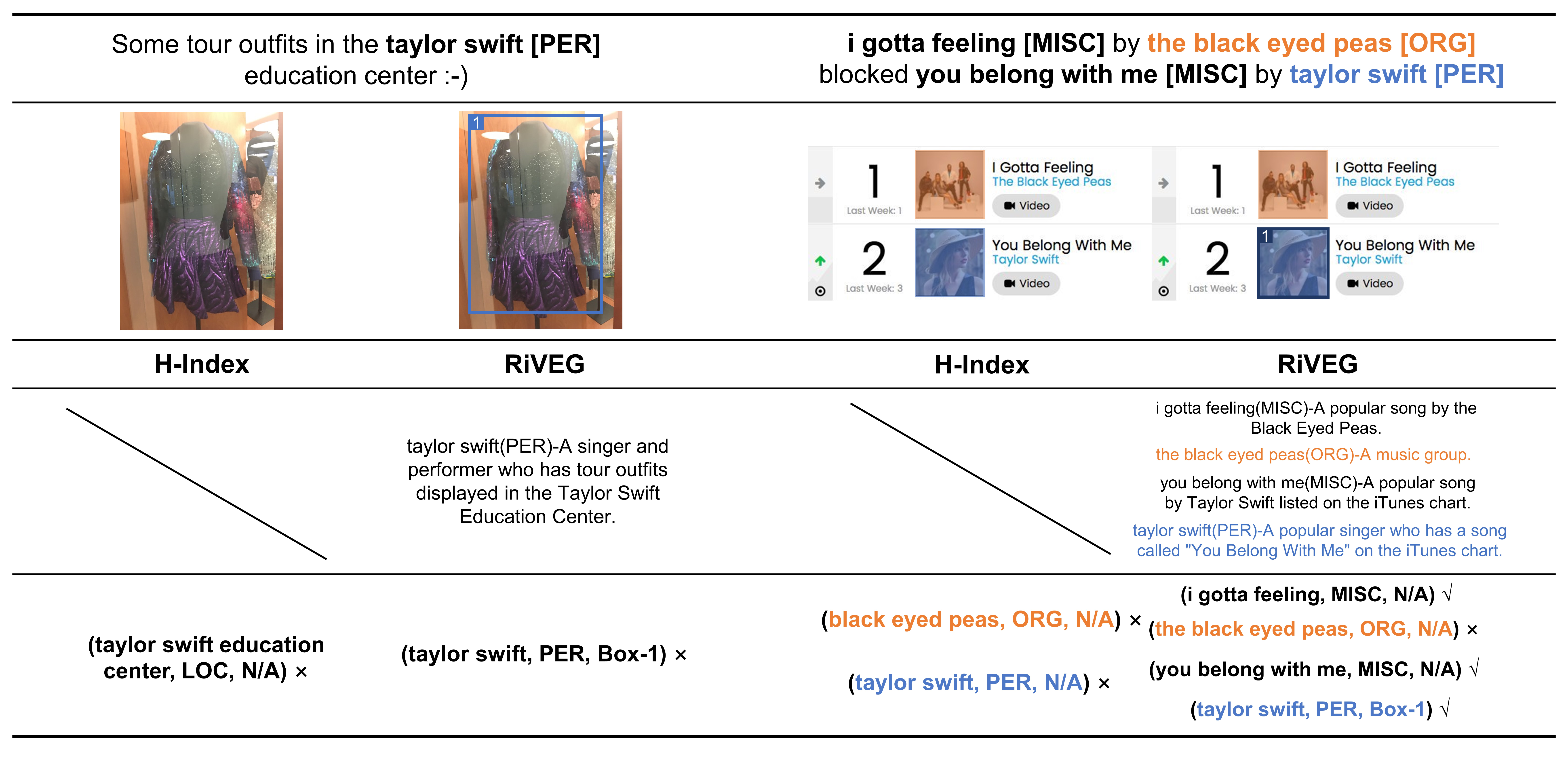}
	\caption{A case study on two incorrect predictions.}
	\label{fig:cases wrong}
\end{figure*}

Figure \ref{fig:cases wrong} illustrates two error examples. The left test sample poses a significant challenge for entity grounding due to the resemblance between the shape of the clothing in the image and that of humans. H-Index fails to predict the named entity. While RiVEG accurately predicts the entity-type pairs, it incorrectly performs entity grounding. The entity prediction and grounding for the sample on the right are both highly challenging. In this case, H-Index is almost incapable of making effective predictions. RiVEG accurately predicts all entity-type pairs and successfully performs the grounding of ``\emph{taylor swift}''. However, it fails to accurately complete the grounding of ``\emph{the black eyed peas}''. This indicates that the GMNER task remains highly challenging. 

\begin{figure*}
    \centering
    \includegraphics[scale=0.8]{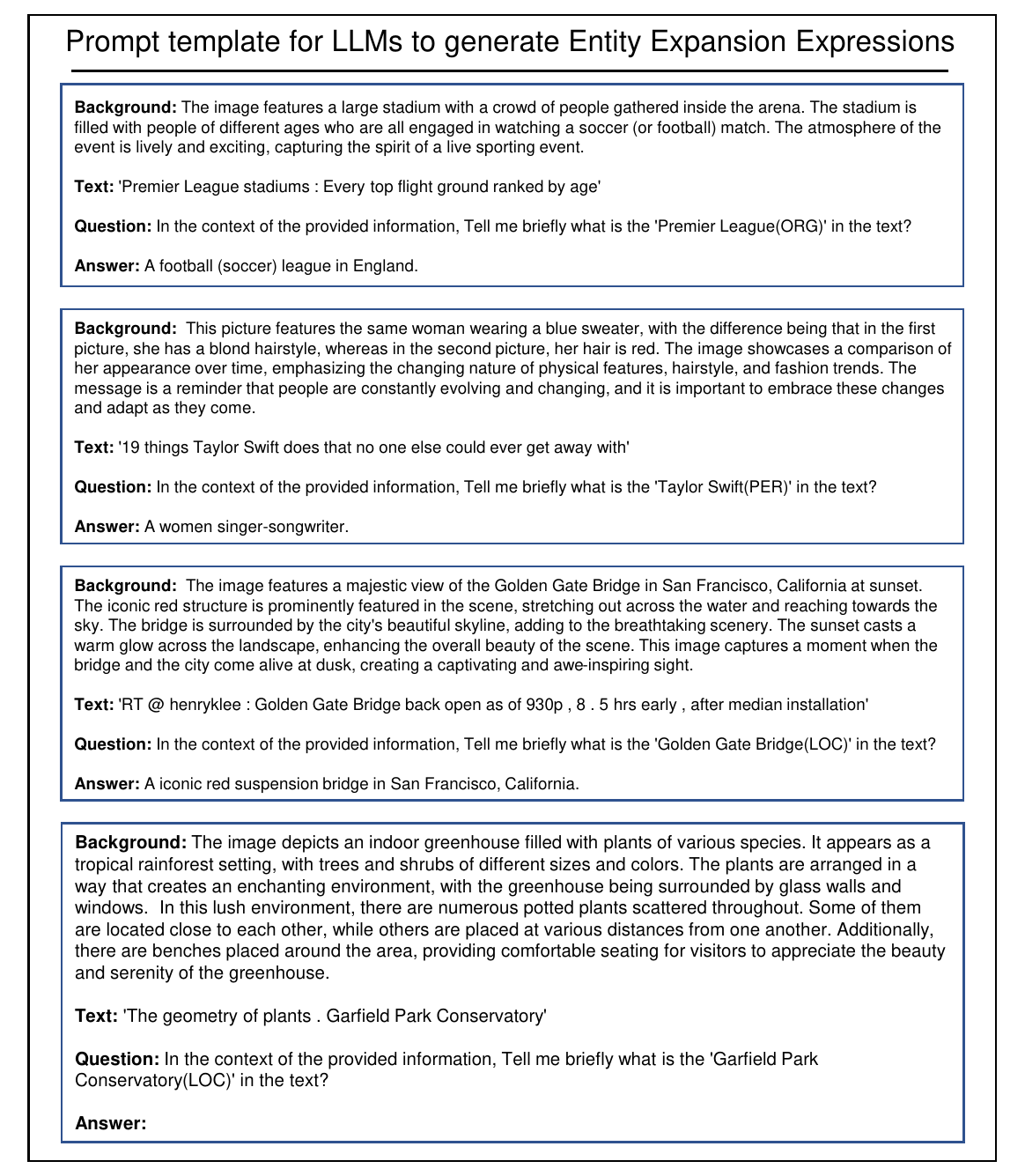}
    \caption{A prompt template for LLMs to generate entity expansion expressions.}
    \label{in-context sample}
\end{figure*}

\end{document}